# Association Rule Pruning based on Interestingness Measures with Clustering

**S.Kannan[1] and R.Bhaskaran[2]**

**[1]Department of Computer Science, DDE, Madurai Kamaraj University
Madurai-625021, Tamil Nadu, India.**

**[2]School of Mathematics, Madurai Kamaraj University
Madurai-625021, Tamil Nadu, India.**

## Abstract

Association rule mining plays vital part in knowledge mining. The difficult task is discovering knowledge or useful rules from the large number of rules generated for reduced support. For pruning or grouping rules, several techniques are used such as rule structure cover methods, informative cover methods, rule clustering, etc. Another way of selecting association rules is based on interestingness measures such as support, confidence, correlation, and so on. In this paper, we study how rule clusters of the pattern $X_i \rightarrow Y$ are distributed over different interestingness measures.

## 1. Introduction

In this world of fast information communication, massive amount of data is generated and stored in computer database systems. Association rule mining (ARM) is the most popular knowledge discovery technique used in several areas of applications. In ARM, large number of Association rules or patterns or knowledge is generated from the large volume of dataset. But most of the association rules have redundant information and thus all of them can not be used directly for an application. So pruning or grouping rules by some means is necessary to get very important rules or knowledge. One way of selecting very interesting rules is using interestingness measures to rank and select a small set of rules of different characteristics. Another way is forming groups or clusters of rules and selecting very important rules from each cluster.

This paper is organized as follows. Section 2 describes about association rules and their interestingness measures. Method of grouping or clustering rules and selection of cover rules are discussed in section 3. Related research works are mentioned in section 4. Section 5 gives details about the dataset used for this work. In section 6, our method of analyzing rule distribution in different clusters for different interestingness measures is discussed. Results, discussion and conclusion are given in section 7.

## 2. Association rules and their interestingness measures

Let $\mathbf{D}$ be a dataset with $|\mathbf{D}|$ instances or tuples. Let $\mathbf{I}=\{I_1, I_2,\ldots, I_m\}$ be set of m distinct attributes or items and each instance in $\mathbf{D}$ is $T \subseteq \mathbf{I}$. For the dataset, association rules[1] of the form $X \rightarrow Y$, where the item-sets $X, Y \subset \mathbf{I}$ and $X \cap Y = \varnothing$, are generated using methods Apriori[2][4], FP-Growth[3][4] or any other well known techniques. The item-sets $X$ and $Y$ are called antecedent and consequent of the rule respectively. Generation of association rules(AR) is generally controlled by the two measures or metrics called support and confidence, which are given below.

Table-1 Weather Dataset

| Record No. | Outlook | Temperature | Humidity | Windy | Play |
|---|---|---|---|---|---|
| 1 | sunny | hot | high | False | no |
| 2 | sunny | hot | high | True | no |
| 3 | overcast | hot | high | False | yes |
| 4 | Rainy | mild | high | False | yes |
| 5 | Rainy | cool | normal | False | yes |
| 6 | Rainy | cool | normal | True | no |
| 7 | overcast | cool | normal | True | yes |
| 8 | sunny | mild | high | False | no |
| 9 | sunny | cool | normal | False | yes |
| 10 | Rainy | mild | normal | False | yes |
| 11 | sunny | mild | normal | True | yes |
| 12 | overcast | mild | high | True | yes |
| 13 | overcast | hot | normal | False | yes |
| 14 | Rainy | mild | high | True | no |

$Support$ = P($X \cup Y$) = P($XY$)

= (Number of tuples that contains both $X$ and $Y$ ) / (Total number of tuples in **D**)

$Confidence$ = P($Y \mid X$) = P($X \cup Y$) / P($X$)

= P($XY$) / P($X$)

For example consider the weather dataset given in table-1. Following are the sample association rules generated for minimum support 20% and minimum confidence 70%. The number of records covered by antecedent and the rule are given in the left side and right side of the implication mark respectively. The records numbers covered as in table-1 are mentioned within parenthesis.

**Sample Rules:**

R1. Outlook=overcast 4(3,7,12,13) ==> Play=yes 4(3,7,12,13)   sup:(29%) conf:(100%)

R2. Temperature=cool 4(5,6,7,9) ==> Humidity=normal 4(5,6,7,9)   sup:(29%) conf:(100%)

R3. Humidity=normal ^ Windy=FALSE 4(5,9,10,13) ==> Play=yes 4(5,9,10,13)   sup:(29%) conf:(100%)

R4. Outlook=sunny ^ Play=no 3(1,2,8) ==> Humidity=high 3(1,2,8)   sup:(21%) conf:(100%)

R5. Outlook=sunny ^ Humidity=high 3(1,2,8) ==> Play=no 3(1,2,8)   sup:(21%) conf:(100%)

R6. Outlook=rainy ^ Play=yes 3(4,5,10) ==> Windy=FALSE 3(4,5,10)   sup:(21%) conf:(100%)

R7. Outlook=rainy ^ Windy=FALSE 3(4,5,10) ==> Play=yes 3(4,5,10)   sup:(21%) conf:(100%)

R8. Temperature=cool ^ Play=yes 3(5,7,9) ==> Humidity=normal 3(5,7,9)   sup:(21%) conf:(100%)

R9. Humidity=normal 7(5,6,7,9,10,11,13) ==> Play=yes 6(5,7,9,10,11,13)   sup:(43%) conf:(86%)

R10. Play=no 5(1,2,6,8,14) ==> Humidity=high 4(1,2,8,14)   sup:(29%) conf:(80%)

R11. Windy=FALSE 8(1,3,4,5,8,9,10,13) ==> Play=yes 6(3,4,5,9,10,13)   sup:(43%) conf:(75%)

R12. Temperature=hot 4(1,2,3,13) ==> Humidity=high 3(1,2,3)   sup:( 21%) conf:(75%)

R13. Temperature=hot 4(1,2,3,13) ==> Windy=FALSE 3(1,3,13)   sup:( 21%) conf:(75%)

R14. Temperature=cool 4(5,6,7,9) ==> Play=yes 3(5,7,9)   sup:( 21%) conf:(75%)

R15. Humidity=high ^ Play=no 4(1,2,8,14) ==> Outlook=sunny 3(1,2,8)   sup:( 21%) conf:(75%)

R16. Temperature=cool ^ Humidity=normal 4(5,6,7,9) ==> Play=yes 3(5,7,9)   sup:( 21%) conf:(75%)

R17. Temperature=cool 4(5,6,7,9) ==> Humidity=normal Play=yes 3(5,7,9)   sup:( 21%) conf:(75%)

Number of rules grows to several thousands if the support and confidence thresholds are reduced to low. To select interesting rules, different interesting measures [5] are used to rank the generated rules. Each measure has its own selection characteristics and its own positives and negatives. For detailed discussion, refer [5][6][7][8]. The table-2 list most generally used interestingness measures with their formula for computation.

# 3. Method of clustering rules and selecting cover rules

Discovered rules with the given confidence and support thresholds are large in number. All these rules are not useful, since they are heavily redundant in information. There are several ways of grouping rules such as methods based on clustering techniques[9], rule structure[10][11][13], rule instance cover[9][10][12] and so on. In our work, rules are grouped based on rule consequent information. So groups of rules are in the form $X_i \rightarrow Y$ for i=1,2,...,n. That is, different rule antecedents $X_i$'s are collected into one group for a same rule consequent $Y$. For our example, from the above rules generated for weather dataset, eight rules R1, R3, R7, R9, R11, R14, R16 and R17 are collected into a group or cluster named as $\mathbf{R}_{Play=yes}$={R1, R3, R7, R9, R11, R14, R16, R17}. Like this, several clusters of rules such as $\mathbf{R}_{Humidity=normal}$, $\mathbf{R}_{Humidity=high}$ and so on can be formed.

Since each group has large number of rules, next step is to select small set of representative rules from each group. Representative rules are selected based on rule instance cover as follows.

Let $\mathbf{R}$y={ $X_i \rightarrow Y$ | i=1,2,...,n } be a set of n rules for some item-set $Y$ and m($X_i$ $Y$) be rule cover, which is the set of tuples/records covered by the rule $X_i \rightarrow Y$ in the dataset **D**. Let $\mathbf{C}_y$ be the cluster rule cover for a group or cluster of rules $\mathbf{R}$y.  i.e.,

$\mathbf{C}_y$ = m($\mathbf{R}$y) = $\cup_{i=1,2,...n}$ m($X_i Y$)

For our example, the cluster cover $\mathbf{C}_{Play=yes}$ for the cluster $\mathbf{R}_{Play=yes}$ can be computed as follows.

$\mathbf{C}_{Play=yes}$ ={m(R1) $\cup$ m(R3) $\cup$ m(R7) $\cup$ m(R9) $\cup$ m(R11) $\cup$ m(R14) $\cup$ m(R16) $\cup$ m(R17)}

= { R1{3,7,12,13} $\cup$ R3{5,9,10,13} $\cup$ R7{4,5,10} $\cup$ R9{5,7,9,10,11,13} $\cup$ R11{3,4,5,9,10,13} $\cup$ R14{5,7,9} $\cup$ R16{5,7,9} $\cup$ R17{5,7,9} }

= $\mathbf{C}_{Play=yes}${3,4,5,7,9,10,11,12,13}

Next, from cluster rule set $\mathbf{R}$y, find a small set of k rules $\mathbf{r}_y$ called representative rule set such that m($\mathbf{r}_y$) is almost equal to m($\mathbf{R}$y). i.e.,

m($\mathbf{r}_y$) $\cong$ m($\mathbf{R}$y),   or

$\cup_{j=1,2,...k}$ m($X_i Y$) $\approx \cup_{i=1,2,...n}$ m($X_i Y$), where k<< n

To find representative rule set $\mathbf{r}_y$ from $\mathbf{R}$y, we use the following rule cover algorithm, which is a modified version of algorithm given in [10].

Table-2: Interestingness Measures used for Association Rules

| Measure | Formula |
|---|---|
| Support | $P(XY)$ |
| Confidence/Precision | $P(Y \mid X)$ |
| Coverage | $P(X)$ |
| Prevalence | $P(Y)$ |
| Recall / Sensitivity | $P(X \mid Y)$ |
| Specificity-1 | $P(\neg Y \mid \neg X)$ |
| Accuracy | $P(XY) + P(\neg X \neg Y)$ |
| Lift/Interest | $P(Y \mid X)/P(Y)$ or $P(XY)/P(X)P(Y)$ |
| Leverage-1 | $P(Y \mid X) - P(X)P(Y)$ |
| Added Value / Change of Support | $P(Y \mid X) - P(Y)$ |
| Relative Risk | $P(Y \mid X)/P(Y \mid \neg X)$ |
| Jaccard | $P(XY)/(P(X) + P(Y) - P(XY))$ |
| Certainty Factor | $(P(Y \mid X) - P(Y))/(1 - P(Y))$ |
| Odds Ratio | $\{P(XY)P(\neg X \neg Y)\}/\{P(X \neg Y)P(\neg XY)\}$ |
| Yule's Q | $\{P(XY)P(\neg X \neg Y) - P(X \neg Y)P(\neg XY)\}/$ $\{P(XY)P(\neg X \neg Y) + P(X \neg Y)P(\neg XY)\}$ |
| Yule's Y | $\{\sqrt{P(XY)P(\neg X \neg Y)} - \sqrt{P(X \neg Y)P(\neg XY)}\}/$ $\{\sqrt{P(XY)P(\neg X \neg Y)} + \sqrt{P(X \neg Y)P(\neg XY)}\}$ |
| Klosgen | $(\sqrt{P(XY)})(P(Y \mid X) - P(Y)),$ $(\sqrt{P(XY)})\max(P(Y \mid X) - P(Y), P(X \mid Y) - P(X))$ |
| Conviction | $(P(X)P(\neg Y)) / P(X \neg Y)$ |
| Collective Strength | $\{ (P(XY)+P(\neg Y \mid \neg X)) / (P(X)P(Y)+P(\neg X)P(\neg Y)) \}*$ $\{ (1-P(X)P(Y)-P(\neg X)P(\neg Y)) / (1-P(XY)-P(\neg Y \mid \neg X)) \}$ |
| Laplace Correction | $(N(XY)+1) / (N(X)+2)$ |
| Gini Index | $P(X)*\{P(Y \mid X)^2 + P(\neg Y \mid X)^2\} + P(\neg X)*\{P(Y \mid \neg X)^2 + P(\neg Y \mid \neg X)^2\} - P(Y)^2 - P(\neg Y)^2$ |
| $\varnothing$−Coefficient (Linear Correlation Coefficient) | $\{P(XY)-P(X)P(Y)\} / \sqrt{\{P(X)P(Y)P(\neg X)P(\neg Y)\}}$ |
| J-Measure | $P(XY) \log( P(Y \mid X) / P(Y) ) + P(X \neg Y) \log( P(\neg Y \mid X) / P(\neg Y) )$ |
| Piatetsky-Shapiro | $P(XY) - P(X)P(Y)$ |
| Cosine | $P(XY) / \sqrt{(P(X)P(Y))}$ |
| Loevinger | $1 - P(X)P(\neg Y) / P(X \neg Y)$ |
| Information Gain | $\log \{P(XY) / ( P(X)P(Y))\}$ |
| Sebag-Schoenauer | $P(XY) / P(X \neg Y)$ |
| Least Contradiction | $\{P(XY)-P(X \neg Y)\} / P(Y)$ |
| Odd Multiplier | $\{P(XY)P(\neg Y)\} / \{P(Y)P(X \neg Y)\}$ |
| Example and Counterexample Rate | $1 - \{P(X \neg Y) / P(XY)\}$ |
| Zhang | $\{P(XY)-P(X)P(Y)\} / \max(P(XY)P(\neg Y), P(Y)P(X \neg Y))$ |
| Correlation | $\{ P(XY)-P(X)P(Y) \} / \{ P(X)P(Y)(1-P(X))(1-P(Y)) \}$ |
| Leverage-2 | $P(XY) - P(X)P(Y)$ |
| Coherence | $P(XY) / (P(X)+P(Y)-P(XY))$ |
| Specificity-2 | $P(\neg X \mid \neg Y)$ |
| All Confidence | $\min( P(X \mid Y), P(Y \mid X) )$ |
| Max Confidence | $\max( P(X \mid Y), P(Y \mid X) )$ |
| Kulczynski | $(P(X \mid Y)+P(Y \mid X))/2$ |

**Algorithm Rule-Cover**

Input:  Set of rules $\mathbf{R}y = \{ X_i \rightarrow Y \mid i=1,2,\dots,n\}$
        Set of matched tuples/records $m(X_iY)$ for all $i \in \{1,2,\dots,n\}$
Output: Representative rule set $\mathbf{r}_y$
Method :
        $\mathbf{r}_y = \varnothing$

$\mathbf{C}_y = \cup_{i=1,2,\dots,n} m(X_iY)$
$S = |\mathbf{C}_y|$
For $i \in \{1,2,\dots,n\}$ do
        $c_i = m(X_iY)$
Endfor
While ( $|\mathbf{C}_y| > 2\%$ of $S$ )
        Sort all rules $X_i \rightarrow Y$ in $\mathbf{R}y$ in

descending order of $|c_i|$
  Take the first rule **r** with highest rule cover
  If ( $|m(\mathbf{r})| \leq 2$ % of S)
    Exit while loop
  Endif
  $\mathbf{r}_y = \mathbf{r}_y \cup \mathbf{r}$
  $\mathbf{C}_y = \mathbf{C}_y \setminus m(\mathbf{r})$
  For all $i \in (1,2,\ldots,n)$
    $c_i = c_i \setminus m(\mathbf{r})$
  Endfor
Endwhile

For our example cluster $\mathbf{R}_{Play=yes}$={R1, R3, R7, R9, R11, R14, R16, R17}, based on the above algorithm rule R9(5,7,9,10,11,13) is selected first, then rule R1(3,12) is selected and finally R7(4) is selected. So the representative rule set is $\mathbf{r}_{Play=yes}$={R9,R1, R7}.

## 4. Related Work

Since there are large number of patterns or association rules(AR) generated in ARM, clustering association rules is one of the meaningful way of grouping related patterns or association rules into different clusters. Association rule coverage is an efficient way of selecting cluster cover representative rules.

In [9], the authors selected highly ranked (based on confidence) association rules one by one and formed cluster of objects covered by each rule until all the objects in the database are covered. The authors of [10] formed cluster of rules of the form $Xi \rightarrow Y$, that is, rules with different antecedent but with same consequent $Y$ and they extracted representative rules for each cluster as knowledge for the cluster. In [11], the authors formed cluster of rules based on structure distance of antecedent. The authors of [12] formed hierarchical clustering of rules based on different distance methods used for rules. In [13], the authors discussed different ways of pruning redundant rules including rule cover method. All Associative Classifier (AC) CBA, CMAR[14], RMR[15], and MCAR[16] generate cluster of rules called class-association rule (CAR) with class label as same consequent and they use database (rule) cover to select potential rules to build (AC) classifier model.

In [5], the authors discussed in detail about classification, properties, characteristics of all interestingness measures used for association rules. In most of the ARM work, confidence measure is used to rank association rules. In associative classifier (AC), confidence, support, and cardinality/size of rule antecedent are used to rank rules. Also, other measures such as chi-square, laplace-accuracy are used to select highly ranked rules.

## 5. Data Source

We are analyzing distance learning program (DLP) student's dataset, which contain details about personnel, school studies, seminar classes, materials used, syllabus and other feedbacks. We collected data randomly through questionnaire from UG and PG students of different courses from different seminar centers of the DLP program. After preprocessing, 2680 samples were used for this analysis. 71 nominal attributes are used in this dataset. Since in this work, we are going to only analyze different clusters, and number of rules covered by each cluster, we do not mention here the details of different attributes and the rules generated.

## 6. Analysis of interestingness measure with the distribution of clusters

To analyze the role of interestingness measures in the distribution of clusters, the following method is used. First, from the dataset the association rules are generated for the given support and confidence thresholds. The association rules of the form $Xi \rightarrow Y$ with the same consequent attributes $Y$ (like Play=yes in our example) but different antecedents attributes $Xi$ are grouped into one cluster. For different consequent attribute set $Y$ (like Humidity=high, Play=no and so on), different cluster of rules is formed. Based on the above rule cover algorithm, for each cluster of rules, a small set of representative rules are discovered.

**Pruning:** Since there are large number of rules present in each cluster, forming clusters and discovering representative rules is a difficult task. Instead of processing all (several thousands of) association rules, a few thousands (say less than 50%) of highly ranked rules may be used to form clusters and to find representative rules of each cluster. Care must be given in fixing percentage for selecting highly ranked rules, since selection of very small percentage of rules may not cover all the clusters effectively. Interestingness measures such as mentioned in table-2 can be used to rank the association rules. The highly ranked small percentage of rules is used to form clusters and to discover representative rules for each cluster. These discovered rules of each cluster are compared with the representative rules that are discovered for the corresponding cluster using the whole set of association rules.

This experiment is repeated for each interestingness measure to rank and prune association rules, to discover representative rules and to compare. We analyze how the representative rules of different clusters from pruned rules coincide or deviate with the representative rules discovered from the whole set of association rules without pruning.

## 7. Experimental Results, Discussion and Conclusion

### 7.1 Results

For our DLP student's real dataset, 2332 association rules are generated for 30% minimum support and 80% minimum confidence thresholds. Using the above mentioned method, 18 clusters (of association rues) for 18 different consequent attribute sets are formed using all 2332 association rules (All-ARs) and for each cluster the representative rules are extracted. Total number of representative rules for each cluster is found for the 18 clusters and are given in the 2nd row of appendix-A as All-ARs under the corresponding cluster number (in 1st row).

For each of the 2332 association rules, all the interestingness measures as given in table-2 are computed. The rules are ordered based on a measure for example Accuracy and the top highly ranked 500 rules (nearly 21%) from the ordered 2332 rules are used to form clusters and discover representative rules for each of the 18 clusters. The total number of rules for each cluster is taken and is given as *Cluster* (in the 3rd row of Appendix-A for Accuracy). The representative rules of each cluster are compared with that of corresponding cluster of All-ARs and the number of common representative rules for each cluster is taken and is entered as *Common* for the 18 clusters (in the 4th row of Appendix-A for Accuracy).

Table–3: Total number of *Cluster* and *Common* representative rules for all interestingness measures

| Interestingness Measures | 36 clusters using 5000 rules from rules generated with Support 20% (refer Appendix B) | | 18 clusters using 500 rules from rules generated with Support 30% (refer Appendix A) | | 18 clusters using 1000 rules from rules generated with Support 30% | |
|---|---|---|---|---|---|---|
| | Total *Cluster* Rules | Total *Common* Rules | Total *Cluster* Rules | Total *Common* Rules | Total *Cluster* Rules | Total *Common* Rules |
| All-Ars | 104 | | 44 | | 44 | |
| Accuracy | 103 | 101 | 40 | 40 | 44 | 44 |
| All-Confidence | 90 | 78 | 29 | 29 | 42 | 42 |
| Certainity | 87 | 77 | 45 | 25 | 45 | 35 |
| Conviction | 116 | 66 | 45 | 25 | 45 | 35 |
| Coherence | 87 | 77 | 28 | 28 | 39 | 37 |
| Confidence | 5 | 3 | 4 | 2 | 7 | 5 |
| Correlation | 102 | 97 | 40 | 36 | 45 | 39 |
| Cosine | 85 | 74 | 27 | 27 | 37 | 36 |
| Col-strength | 116 | 86 | 46 | 31 | 47 | 32 |
| Changl-Support | 113 | 91 | 51 | 32 | 45 | 37 |
| Coverage | 60 | 49 | 23 | 22 | 30 | 29 |
| Exam-Cex-rate | 5 | 3 | 4 | 2 | 7 | 5 |
| Gini-index | 100 | 98 | 41 | 41 | 41 | 41 |
| Infor-gain | 106 | 91 | 48 | 32 | 46 | 38 |
| Jacard | 87 | 77 | 28 | 28 | 39 | 37 |
| J-measure | 101 | 95 | 41 | 35 | 45 | 38 |
| Klosgen | 101 | 96 | 41 | 36 | 45 | 38 |
| Kulc | 72 | 64 | 22 | 22 | 29 | 29 |
| Least-contraction | 85 | 74 | 24 | 24 | 33 | 32 |
| Linear-Correlation | 102 | 97 | 40 | 36 | 45 | 39 |
| Lift-Interest | 106 | 91 | 48 | 32 | 46 | 38 |
| Loevinger | 14 | 4 | 14 | 2 | 13 | 5 |
| Laplace-Correction | 5 | 3 | 4 | 2 | 7 | 5 |
| Leverage | 73 | 14 | 15 | 3 | 47 | 7 |
| Leverage-2 | 105 | 97 | 39 | 37 | 46 | 38 |
| Max-confidence | 6 | 4 | 5 | 3 | 7 | 6 |
| Odd-Multiplier | 113 | 71 | 44 | 30 | 44 | 37 |
| Odds-Ratio | 98 | 81 | 39 | 35 | 44 | 38 |
| Piatetsky-Shapiro | 105 | 97 | 39 | 37 | 46 | 38 |
| Prevalance | 3 | 1 | 3 | 1 | 6 | 3 |
| Qyule | 97 | 81 | 39 | 35 | 44 | 38 |
| Recall/Sensitivity | 90 | 78 | 29 | 29 | 42 | 42 |
| Relative-Risk | 100 | 97 | 40 | 37 | 45 | 40 |
| Sebag-Schoenauer | 5 | 3 | 4 | 2 | 7 | 5 |
| Specificity | 101 | 100 | 42 | 40 | 41 | 41 |
| Specificity-2 | 137 | 44 | 53 | 14 | 59 | 28 |
| Support | 53 | 44 | 17 | 17 | 28 | 28 |
| Yyule | 97 | 81 | 39 | 35 | 44 | 38 |
| Zhang | 113 | 71 | 44 | 30 | 44 | 37 |

This process is repeated for each other interestingness measures (such as All-Confidence, Certainity, and so on as given in table-2) and the total number of *Cluster* and *Common* representative rules for each of all the 18 clusters are computed and are given in Appendix-A. The total *Cluster* and *Common* rules are given in the last column of Appendix-A. These details are given in the fourth and fifth columns of table-3 as Total *Cluster* and Total *Common* rules respectively.

The above process is repeated for all interesting measures with top 1000 rules (nearly 43%) instead of 500 rules. Total number of *Cluster* and *Common* representative rules for all measures are computed and entered in the sixth and seventh columns of table-3.

For the same dataset, 23417 association rules are generated for 20% minimum support and 80% minimum confidence. The above experiment is performed using top 5000 rules (nearly 21%) based on each of all interesting measure and the total number of *Cluster* and *Common* representative rules for all measures are computed and entered in the second and third columns of table-3.

## 7.2 Discussion

**Analysis for the same level of Pruning:** Following are the observations for the same level of pruning based on different measures. The measures which cover considerable representative rules are taken for comparative analysis. In Figure-1, only for these measures, the total *cluster* and *common* rules are plotted in the decreasing order of total *cluster* rules. The results using the same percentage (i.e., 21%) of rules from the rules generated with 30% support and 20% support are plotted in Figures 1a (using 500 rules) and 1b (using 5000 rules) respectively. When compared the *cluster* and *common* representative rules for each measure for 30% support with that of 20% support from figures 1a and 1b, most of the measures give almost similar results for both 30% support and 20% support.

The measures specificity-2, conviction, zhang and odd-multiplier cover more and different rules when compared with the rules covered by All-ARs. In particular, specificity-2 covers large number of rules than other measures. Also, the measures change-of-support, collective-strength, information-gain, and lift/interest cover more rules. But the *common* rules are more and so they are not very different from All-ARs. The cover rules by the measures accuracy, specificity, gini-index,

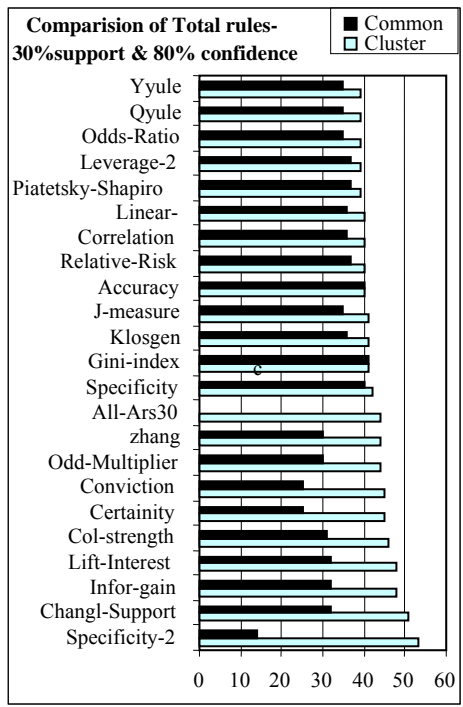

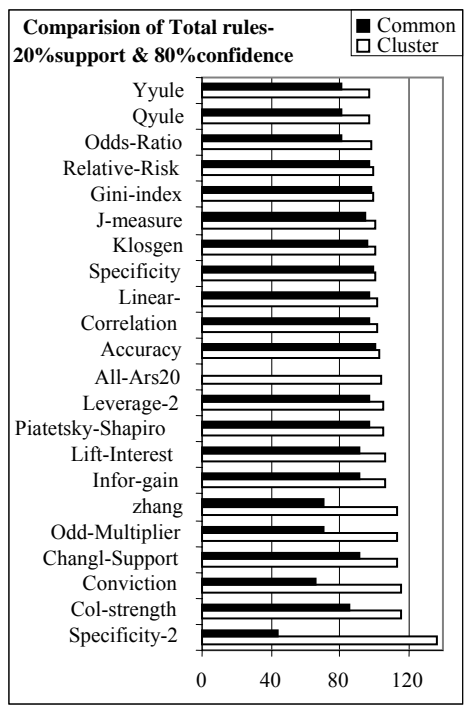

Figure 1a

Figure 1b

Figure-1 Comparison of total *cluster* and *common* representative rules using (a) Top 500 Rules generated with 30% support and 80% confidence and (b) Top 5000 Rules generated with 20% support and 80% confidence

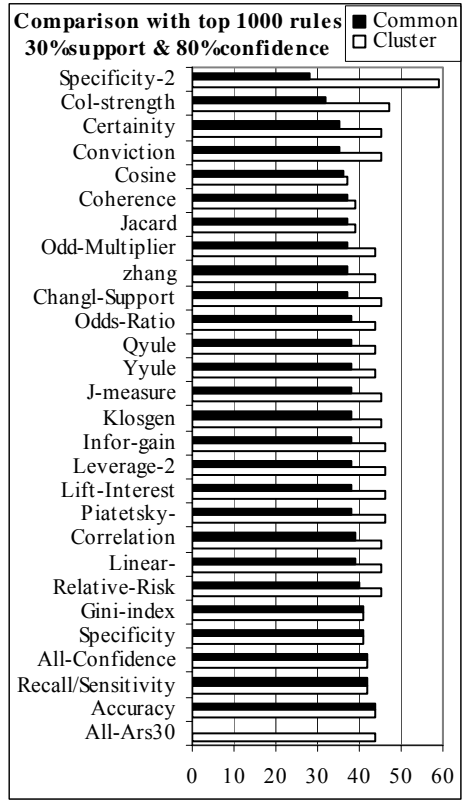

Figure 2a

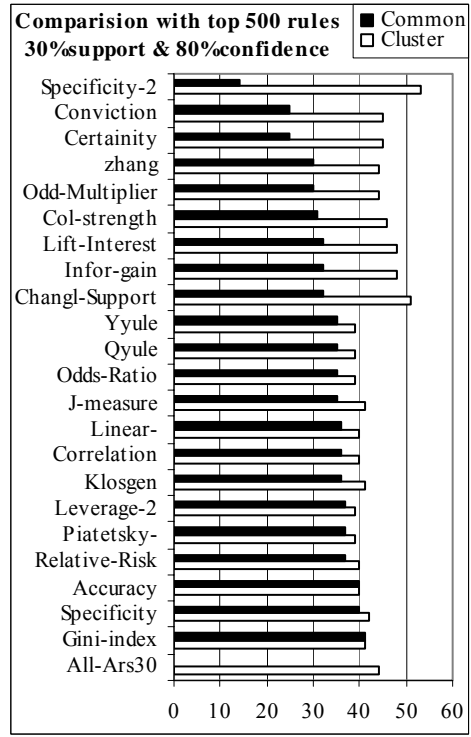

Figure 2b

Figure-2 Comparison of total *cluster* and *common* representative rules using (a) Top 1000 rules and (b) Top 500 rules from Rules generated with 30% support and 80% confidence

correlation, linear-correlation, piatetsky-shapiro, relative-risk, linear-correlation, klosgen, j-measure and leverage-2 are similar to All-ARs (i.e., most *cluster* rules are *common* to All-ARs). In particular, for the measures accuracy, specificity, and gini-index, they cover almost all the rules of All-ARs.

**Analysis for the different level of Pruning:** Following are the observations made on cluster rules for different level of pruning based on interestingness measures. The results of total *cluster* and *common* rules by using top 1000 rules (43% of rules) and top 500 rules (21% of rules) from the 2332 rules generated with 30% support and 80% confidence are plotted respectively in figures 2(a) and 2(b) in the decreasing order of total *common* rules. When compared, almost all measures give similar results for both top 1000 rules and top 500 rules. But total *cluster* and *common* rules are slightly more for using 1000 rules.

Here also, the measures specificity-2, conviction, certainity, zhang, odd-multiplier, col-strength, lift/interest, infor-gain, and change-of-support give large number of *cluster* representative rules most of which are different from All-ARs. But the *common* rules are increasing more for the use of more top (ranked) rules. Also, the measures accuracy, specificity, and gini-index give representative rules almost similar to All-ARs (i.e. almost all are common representative rules). This coincides with the results for the same level of pruning.

In addition for the increase of number of top rules, some other measures such as cosine, coherence, jacard, all-confidence, and recall/sensitivity give considerable number of representative rules. Particularly, all-confidence, and recall/sensitivity give better results.

## 8. Conclusion

Since the measures accuracy, specificity and gini-index give similar results as that of using all association rules, they can be used to prune rules. After pruning, only less than 50% of the total rules need be used to extract cluster representative rules without any loss of information or knowledge. Also to extract different set of more representative rules, specificity-2 can be used. This gives different type of information or knowledge for the same clusters.

As future work, we intend to study the effect of this pruning with the efficiency of classification based on association rule (CBA) model. Also, we intend to use the above measures in ranking Class Association Rules (CARs).

Appendix -A

| | Cluster-Number → | 1 | 2 | 3 | 4 | 5 | 6 | 7 | 8 | 9 | 10 | 11 | 12 | 13 | 14 | 15 | 16 | 17 | 18 | Total Rules |
|---|---|---|---|---|---|---|---|---|---|---|---|---|---|---|---|---|---|---|---|---|
| All-Ars | Cluster | 3 | 1 | 4 | 4 | 1 | 1 | 1 | 1 | 1 | 3 | 1 | 3 | 5 | 1 | 6 | 3 | 3 | 2 | 44 |
| | Cluster | 1 | 1 | 4 | 3 | 1 | 1 | 1 | 1 | 1 | 3 | 1 | 3 | 5 | 1 | 5 | 3 | 3 | 2 | 40 |
| Accuracy | Common | 1 | 1 | 4 | 3 | 1 | 1 | 1 | 1 | 1 | 3 | 1 | 3 | 5 | 1 | 5 | 3 | 3 | 2 | 40 |
| All-Confidence | Cluster | 1 | 1 | 1 | 2 | 1 | 1 | 1 | 1 | 1 | 2 | 1 | 3 | 2 | 1 | 2 | 3 | 3 | 2 | 29 |
| | Common | 1 | 1 | 1 | 2 | 1 | 1 | 1 | 1 | 1 | 2 | 1 | 3 | 2 | 1 | 2 | 3 | 3 | 2 | 29 |
| Certainty | Cluster | 3 | 1 | 4 | 3 | 1 | 1 | 1 | 1 | 1 | 3 | 1 | 4 | 4 | 1 | 4 | 4 | 4 | 4 | 45 |
| | Common | 3 | 1 | 4 | 0 | 1 | 1 | 1 | 1 | 1 | 3 | 1 | 0 | 4 | 1 | 3 | 0 | 0 | 0 | 25 |
| Conviction | Cluster | 3 | 1 | 4 | 3 | 1 | 1 | 1 | 1 | 1 | 3 | 1 | 4 | 4 | 1 | 4 | 4 | 4 | 4 | 45 |
| | Common | 3 | 1 | 4 | 0 | 1 | 1 | 1 | 1 | 1 | 3 | 1 | 0 | 4 | 1 | 3 | 0 | 0 | 0 | 25 |
| Coherence | Cluster | 1 | 1 | 1 | 1 | 1 | 1 | 1 | 1 | 1 | 2 | 1 | 3 | 2 | 1 | 2 | 3 | 3 | 2 | 28 |
| | Common | 1 | 1 | 1 | 1 | 1 | 1 | 1 | 1 | 1 | 2 | 1 | 3 | 2 | 1 | 2 | 3 | 3 | 2 | 28 |
| Confidence | Cluster | 0 | 0 | 0 | 0 | 0 | 0 | 1 | 0 | 0 | 0 | 0 | 3 | 0 | 0 | 0 | 0 | 0 | 0 | 4 |
| | Common | 0 | 0 | 0 | 0 | 0 | 0 | 1 | 0 | 0 | 0 | 0 | 1 | 0 | 0 | 0 | 0 | 0 | 0 | 2 |
| Correlation | Cluster | 3 | 1 | 4 | 2 | 1 | 1 | 1 | 1 | 1 | 3 | 1 | 3 | 5 | 1 | 6 | 2 | 2 | 2 | 40 |
| | Common | 3 | 1 | 4 | 2 | 1 | 1 | 1 | 1 | 1 | 3 | 1 | 0 | 5 | 1 | 6 | 2 | 2 | 1 | 36 |
| Cosine | Cluster | 1 | 1 | 1 | 1 | 1 | 1 | 1 | 1 | 1 | 1 | 1 | 3 | 2 | 1 | 2 | 3 | 3 | 2 | 27 |
| | Common | 1 | 1 | 1 | 1 | 1 | 1 | 1 | 1 | 1 | 1 | 1 | 3 | 2 | 1 | 2 | 3 | 3 | 2 | 27 |
| Col-strength | Cluster | 3 | 3 | 4 | 3 | 1 | 3 | 0 | 0 | 0 | 3 | 2 | 3 | 5 | 1 | 6 | 3 | 3 | 3 | 46 |
| | Common | 3 | 0 | 4 | 3 | 1 | 0 | 0 | 0 | 0 | 3 | 0 | 1 | 5 | 1 | 6 | 2 | 1 | 1 | 31 |
| Changl-Support | Cluster | 3 | 1 | 4 | 4 | 1 | 1 | 1 | 1 | 1 | 3 | 1 | 3 | 5 | 1 | 6 | 5 | 5 | 5 | 51 |
| | Common | 3 | 1 | 4 | 3 | 1 | 1 | 1 | 1 | 1 | 3 | 1 | 0 | 5 | 1 | 6 | 0 | 0 | 0 | 32 |
| Coverage | Cluster | 1 | 1 | 1 | 2 | 1 | 1 | 0 | 0 | 1 | 0 | 1 | 3 | 1 | 1 | 1 | 3 | 3 | 2 | 23 |
| | Common | 1 | 1 | 1 | 2 | 1 | 1 | 0 | 0 | 1 | 0 | 1 | 2 | 1 | 1 | 1 | 3 | 3 | 2 | 22 |
| Exam-Cex-rate | Cluster | 0 | 0 | 0 | 0 | 0 | 0 | 1 | 0 | 0 | 0 | 0 | 3 | 0 | 0 | 0 | 0 | 0 | 0 | 4 |
| | Common | 0 | 0 | 0 | 0 | 0 | 0 | 1 | 0 | 0 | 0 | 0 | 1 | 0 | 0 | 0 | 0 | 0 | 0 | 2 |
| Gini-index | Cluster | 3 | 1 | 4 | 4 | 1 | 1 | 1 | 1 | 1 | 3 | 1 | 0 | 5 | 1 | 6 | 3 | 3 | 2 | 41 |
| | Common | 3 | 1 | 4 | 4 | 1 | 1 | 1 | 1 | 1 | 3 | 1 | 0 | 5 | 1 | 6 | 3 | 3 | 2 | 41 |
| Infor-gain | Cluster | 3 | 1 | 4 | 4 | 1 | 1 | 1 | 1 | 1 | 3 | 1 | 0 | 5 | 1 | 6 | 5 | 5 | 5 | 48 |
| | Common | 3 | 1 | 4 | 3 | 1 | 1 | 1 | 1 | 1 | 3 | 1 | 0 | 5 | 1 | 6 | 0 | 0 | 0 | 32 |
| Jacard | Cluster | 1 | 1 | 1 | 1 | 1 | 1 | 1 | 1 | 1 | 2 | 1 | 3 | 2 | 1 | 2 | 3 | 3 | 2 | 28 |
| | Common | 1 | 1 | 1 | 1 | 1 | 1 | 1 | 1 | 1 | 2 | 1 | 3 | 2 | 1 | 2 | 3 | 3 | 2 | 28 |
| J-measure | Cluster | 3 | 1 | 4 | 2 | 1 | 1 | 1 | 1 | 1 | 3 | 1 | 4 | 5 | 1 | 6 | 2 | 2 | 2 | 41 |
| | Common | 3 | 1 | 4 | 2 | 1 | 1 | 1 | 1 | 1 | 3 | 1 | 0 | 5 | 1 | 6 | 1 | 2 | 1 | 35 |
| Klosgen | Cluster | 3 | 1 | 4 | 3 | 1 | 1 | 1 | 1 | 1 | 3 | 1 | 2 | 5 | 1 | 6 | 3 | 2 | 2 | 41 |
| | Common | 3 | 1 | 4 | 3 | 1 | 1 | 1 | 1 | 1 | 3 | 1 | 0 | 5 | 1 | 6 | 1 | 2 | 1 | 36 |
| Kulc | Cluster | 1 | 1 | 1 | 1 | 1 | 1 | 1 | 1 | 1 | 1 | 1 | 3 | 0 | 0 | 0 | 3 | 3 | 2 | 22 |
| | Common | 1 | 1 | 1 | 1 | 1 | 1 | 1 | 1 | 1 | 1 | 1 | 3 | 0 | 0 | 0 | 3 | 3 | 2 | 22 |
| Least-contraction | Cluster | 1 | 1 | 1 | 1 | 1 | 1 | 1 | 1 | 1 | 1 | 1 | 3 | 1 | 0 | 1 | 3 | 3 | 2 | 24 |
| | Common | 1 | 1 | 1 | 1 | 1 | 1 | 1 | 1 | 1 | 1 | 1 | 3 | 1 | 0 | 1 | 3 | 3 | 2 | 24 |
| Linear-Correlation | Cluster | 3 | 1 | 4 | 2 | 1 | 1 | 1 | 1 | 1 | 3 | 1 | 3 | 5 | 1 | 6 | 2 | 2 | 2 | 40 |
| | Common | 3 | 1 | 4 | 2 | 1 | 1 | 1 | 1 | 1 | 3 | 1 | 0 | 5 | 1 | 6 | 2 | 2 | 1 | 36 |
| Lift-Interest | Cluster | 3 | 1 | 4 | 4 | 1 | 1 | 1 | 1 | 1 | 3 | 1 | 0 | 5 | 1 | 6 | 5 | 5 | 5 | 48 |
| | Common | 3 | 1 | 4 | 3 | 1 | 1 | 1 | 1 | 1 | 3 | 1 | 0 | 5 | 1 | 6 | 0 | 0 | 0 | 32 |
| Loevinger | Cluster | 0 | 0 | 0 | 0 | 0 | 0 | 0 | 0 | 0 | 0 | 0 | 3 | 0 | 0 | 0 | 3 | 4 | 4 | 14 |
| | Common | 0 | 0 | 0 | 0 | 0 | 0 | 0 | 0 | 0 | 0 | 0 | 1 | 0 | 0 | 0 | 0 | 1 | 0 | 2 |
| Laplace-Correction | Cluster | 0 | 0 | 0 | 0 | 0 | 0 | 1 | 0 | 0 | 0 | 0 | 3 | 0 | 0 | 0 | 0 | 0 | 0 | 4 |
| | Common | 0 | 0 | 0 | 0 | 0 | 0 | 1 | 0 | 0 | 0 | 0 | 1 | 0 | 0 | 0 | 0 | 0 | 0 | 2 |
| Leverage | Cluster | 0 | 1 | 1 | 0 | 0 | 0 | 1 | 1 | 1 | 1 | 2 | 5 | 0 | 0 | 0 | 0 | 2 | 0 | 15 |
| | Common | 0 | 0 | 0 | 0 | 0 | 0 | 1 | 1 | 1 | 0 | 0 | 0 | 0 | 0 | 0 | 0 | 0 | 0 | 3 |
| Max-confidence | Cluster | 0 | 0 | 0 | 0 | 0 | 0 | 1 | 0 | 1 | 0 | 0 | 3 | 0 | 0 | 0 | 0 | 0 | 0 | 5 |
| | Common | 0 | 0 | 0 | 0 | 0 | 0 | 1 | 0 | 1 | 0 | 0 | 1 | 0 | 0 | 0 | 0 | 0 | 0 | 3 |
| Odd-Multiplier | Cluster | 3 | 1 | 4 | 1 | 1 | 1 | 1 | 1 | 1 | 3 | 1 | 4 | 5 | 1 | 6 | 4 | 3 | 3 | 44 |
| | Common | 3 | 1 | 4 | 1 | 1 | 1 | 1 | 1 | 1 | 3 | 1 | 0 | 5 | 1 | 6 | 0 | 0 | 0 | 30 |
| Odds-Ratio | Cluster | 3 | 1 | 4 | 1 | 1 | 1 | 1 | 1 | 1 | 3 | 1 | 3 | 5 | 1 | 6 | 2 | 2 | 2 | 39 |
| | Common | 3 | 1 | 4 | 1 | 1 | 1 | 1 | 1 | 1 | 3 | 1 | 0 | 5 | 1 | 6 | 2 | 2 | 1 | 35 |
| Piatetsky-Shapiro | Cluster | 3 | 1 | 4 | 3 | 1 | 1 | 1 | 1 | 1 | 3 | 1 | 1 | 5 | 1 | 6 | 2 | 2 | 2 | 39 |
| | Common | 3 | 1 | 4 | 3 | 1 | 1 | 1 | 1 | 1 | 3 | 1 | 0 | 5 | 1 | 6 | 2 | 2 | 1 | 37 |
| Prevalance | Cluster | 0 | 0 | 0 | 0 | 0 | 0 | 0 | 0 | 0 | 0 | 0 | 3 | 0 | 0 | 0 | 0 | 0 | 0 | 3 |
| | Common | 0 | 0 | 0 | 0 | 0 | 0 | 0 | 0 | 0 | 0 | 0 | 1 | 0 | 0 | 0 | 0 | 0 | 0 | 1 |
| Qyule | Cluster | 3 | 1 | 4 | 1 | 1 | 1 | 1 | 1 | 1 | 3 | 1 | 3 | 5 | 1 | 6 | 2 | 2 | 2 | 39 |
| | Common | 3 | 1 | 4 | 1 | 1 | 1 | 1 | 1 | 1 | 3 | 1 | 0 | 5 | 1 | 6 | 2 | 2 | 1 | 35 |
| Recall/Sensitivity | Cluster | 1 | 1 | 1 | 2 | 1 | 1 | 1 | 1 | 1 | 2 | 1 | 3 | 2 | 1 | 2 | 3 | 3 | 2 | 29 |
| | Common | 1 | 1 | 1 | 2 | 1 | 1 | 1 | 1 | 1 | 2 | 1 | 3 | 2 | 1 | 2 | 3 | 3 | 2 | 29 |

| Measure | Type | | | | | | | | | | | | | | | | | | | Total |
|---|---|---|---|---|---|---|---|---|---|---|---|---|---|---|---|---|---|---|---|---|
| Relative-Risk | Cluster | 3 | 1 | 4 | 3 | 1 | 1 | 1 | 1 | 1 | 3 | 1 | 1 | 5 | 1 | 6 | 2 | 3 | 2 | 40 |
| | Common | 3 | 1 | 4 | 3 | 1 | 1 | 1 | 1 | 1 | 3 | 1 | 0 | 5 | 1 | 6 | 2 | 2 | 1 | 37 |
| Sebag-Schoenauer | Cluster | 0 | 0 | 0 | 0 | 0 | 0 | 1 | 0 | 0 | 0 | 0 | 3 | 0 | 0 | 0 | 0 | 0 | 0 | 4 |
| | Common | 0 | 0 | 0 | 0 | 0 | 0 | 1 | 0 | 0 | 0 | 0 | 1 | 0 | 0 | 0 | 0 | 0 | 0 | 2 |
| Specificity | Cluster | 3 | 1 | 4 | 4 | 1 | 1 | 1 | 1 | 1 | 3 | 1 | 0 | 5 | 1 | 6 | 3 | 3 | 3 | 42 |
| | Common | 3 | 1 | 4 | 4 | 1 | 1 | 1 | 1 | 1 | 3 | 1 | 0 | 5 | 1 | 6 | 3 | 2 | 2 | 40 |
| Specificity-2 | Cluster | 4 | 2 | 4 | 4 | 2 | 2 | 1 | 1 | 1 | 3 | 1 | 5 | 3 | 0 | 1 | 7 | 6 | 6 | 53 |
| | Common | 2 | 0 | 3 | 0 | 0 | 0 | 1 | 1 | 1 | 3 | 1 | 0 | 2 | 0 | 0 | 0 | 0 | 0 | 14 |
| Support | Cluster | 1 | 1 | 1 | 1 | 1 | 1 | 0 | 0 | 0 | 0 | 0 | 3 | 0 | 0 | 0 | 3 | 3 | 2 | 17 |
| | Common | 1 | 1 | 1 | 1 | 1 | 1 | 0 | 0 | 0 | 0 | 0 | 3 | 0 | 0 | 0 | 3 | 3 | 2 | 17 |
| Yyule | Cluster | 3 | 1 | 4 | 1 | 1 | 1 | 1 | 1 | 1 | 3 | 1 | 3 | 5 | 1 | 6 | 2 | 2 | 2 | 39 |
| | Common | 3 | 1 | 4 | 1 | 1 | 1 | 1 | 1 | 1 | 3 | 1 | 0 | 5 | 1 | 6 | 2 | 2 | 1 | 35 |
| zhang | Cluster | 3 | 1 | 4 | 1 | 1 | 1 | 1 | 1 | 1 | 3 | 1 | 4 | 5 | 1 | 6 | 4 | 3 | 3 | 44 |
| | Common | 3 | 1 | 4 | 1 | 1 | 1 | 1 | 1 | 1 | 3 | 1 | 0 | 5 | 1 | 6 | 0 | 0 | 0 | 30 |
| Leverage-2 | Cluster | 3 | 1 | 4 | 3 | 1 | 1 | 1 | 1 | 1 | 3 | 1 | 1 | 5 | 1 | 6 | 2 | 2 | 2 | 39 |
| | Common | 3 | 1 | 4 | 3 | 1 | 1 | 1 | 1 | 1 | 3 | 1 | 0 | 5 | 1 | 6 | 2 | 2 | 1 | 37 |

## Reference:

[1] R. Agrawal, T. Imielinski, and A. Swami, "Mining association rules between sets of items in large databases", In Proc. SIGMOD'93, Washington. DC, May 1993, pp 207-216.

[2] R.Agrawal and R.Srikant, "Fast Algorithms for Mining Association Rules in Large Databases", Proceedings of the 20th International Conference on Very Large Data Bases (VLDB'94), Morgan Kaufmann, 1994, pp 478-499.

[3] J. Han, J. Pei, and Y. Yin. "Mining frequent patterns without candidate generation", In Proc. SIGMOD'00, Dallas, TX, May 2000, pp 1-12.

[4] J.Han and M.Kamber, Data mining: Concepts and Techniques, Morgan Kaufmann, 2001.

[5] Liqiang Geng and Howard J. Hamilton, "Interestingness Measures for Data Mining: A Survey", ACM Computing Surveys, Vol. 38, No. 3, Article 9, September 2006.

[6] P. Tan, V. Kumar, and J. Srivastava. "Selecting the Right Interestingness Measure for Association Patterns". Technical Report 2002-112, Army High Performance Computing Research Center, 2002.

[7] Liaquat Majeed Sheikh, Basit Tanveer, Syed Mustafa Ali Hamdani. "Interesting Measures for Mining Association Rules", In Proceedings of INMIC 2004. 8th International Multitopic Conference, 2004, pp 641-644.

[8]. Tianyi Wu, Yuguo Chen, and Jiawei Han, "Association Mining in Large Databases: A Re-Examination of Its Measures", In Proceedings of PKDD-2007, 11th European Conference on Principles and Practice of Knowledge Discovery in Databases (PKDD), Warsaw, Poland, September 17-21, 2007, pp 621-628.

[9] Waler A.Kosters, Elena Marchiori and Ard A.J.Oerlemans, "Mining Clusters with Association Rules", Advances in Intelligent Data Analysis (IDA-99) (D.J.Hand, J.N.Kok and M.R.Berthold, Eds.), Lecture Notes in Computer Science 1642, Springer, 1999, pp. 39-50.

[10] H. Toivonen, M. Klemettinen, P. Ronkainen, K. Hatonen, and H. Mannila, "Pruning and grouping discovered association rules", In Proc. ECML-95 Workshop on Statistics, Machine Learning, and Knowledge Discovery in Database, April 1995, pp 47-52.

[11] Pi Dechang and Qin Xiaolin, "A new Fuzzy Clustering algorithm on Association rules for knowledge management", Information Technology Journal 7(1), 2008, pp. 119-124.

[12] Alipio Jorge, "Hierarchical Clustering for thematic browsing and summarization of large sets of Association Rules", In Proceedings of the 4th SIAM International Conference on Data Mining. Orlando, FL, 2004, pp. 178-187.

[13] G. Li, and H.J.Hamilton, "Basic association rules", In Proceedings of the 4th SIAM International Conference on Data Mining, Orlando, FL, 2004, pp. 166–177.

[14] W. Li, J. Han, and J. Pei, "CMAR: accurate and efficient classification based on multiple class-association rules", In: Proceedings of IEEE International Conference on Data Mining (ICDM2001), 2001. pp. 369–376

[15] A. Thabtah, and P. I. Cowling, "A greedy classification algorithm based on association rule", Appl. Soft Comput., Vol. 7, No. 3. June 2007, pp. 1102-1111.

[16] Adriano Veloso, Wagner Meira, Marcos Gonçalves, and Mohammed Zaki, "Multi-label Lazy Associative Classification", Knowledge Discovery in Databases: PKDD 2007, 2007, pp. 605-612.

**S.Kannan** He has completed his M.Sc.(physics), M.Sc.(Computer science) and M.Phil.(Comp. Sc.) through Madurai Kamaraj University and M.S.Univerisity (Thirunelveli). Now he is working as Associate Professor in Madurai Kamaraj Univesity. He has presented 4 papers in national and international conferences.

**R.Bhaskaran** He did his M.Sc. from IIT,Chennai in 1973 and obtained his Ph.D. from University of Madras in 1980. He joined the School of Mathematics, Madurai Kamaraj University in 1980. Now he is working as Senior Professor in the School of Mathematics. At present 10 students are working in Data Mining, Image Processing, Software Development, and Character Recognition.